\documentclass[conference]{IEEEtran}
\IEEEoverridecommandlockouts

\def\BibTeX{{\rm B\kern-.05em{\sc i\kern-.025em b}\kern-.08em
    T\kern-.1667em\lower.7ex\hbox{E}\kern-.125emX}}

\usepackage[cmex10]{amsmath}
\usepackage[utf8]{inputenc}
\usepackage{graphicx}
\usepackage{parskip}
\usepackage{ifthen}
\usepackage[dvipsnames]{xcolor}
\usepackage{newtxtext,newtxmath}
\usepackage{float}
\usepackage{makecell}
\usepackage{anyfontsize}

\makeatletter
\let\IEEEcaption\@makecaption
\makeatother

\usepackage[font=footnotesize]{subcaption}
\makeatletter
\let\@makecaption\IEEEcaption
\makeatother

\newcommand*\mean[1]{\overline{#1}}

\usepackage{tikz}
\usetikzlibrary{matrix,positioning,calc,intersections,decorations,chains,decorations.pathreplacing,tikzmark}

\tikzset{LayerNode/.style={
		draw,
		outer sep=0pt,
		inner sep=0.3em,
		on chain}
}

\tikzset{MissingNode/.style={
		draw=none,
		outer sep=0pt,
		inner sep=0.3em,
		on chain}
}

\tikzset{Node/.style={
		draw,
		shape=circle,
		minimum size=2mm,
		inner sep=1pt,
		smooth,
		font=\tiny
	}
}

\tikzset{Link/.style={
		smooth,
		->
	}
}

\newcommand{\filter}[8]{%

	\pgfmathsetmacro\shiftx{#1 / #3 + #4}
	\pgfmathsetmacro\shifty{#2 / #3 + #5}

	\slantedscope{\shiftx}{0}{-\shifty}{0}{
		\fill[White,fill opacity=0.75] (-#6/2,-#7/2) rectangle (#6/2,#7/2);
		\draw[black!75,ultra thin] (-#6/2,-#7/2) rectangle (#6/2,#7/2);
		\node[draw,White,scale=0.05] (l#8n#1) at (0, 0) {};
	}
}

\newcommand{\patch}[6]
{
	\draw[ultra thin,fill=#6] (#1,-#2) rectangle ++(#3,-#3);
	\draw[ultra thin,Black!75] (#1,-#2) -- (#4,-#5);
	\draw[ultra thin,Black!75] (#1 + #3,-#2 - #3) -- (#4,-#5);
	\draw[ultra thin,Black!75] (#1 + #3,-#2) -- (#4,-#5);
	\draw[ultra thin,Black!75] (#1,-#2 - #3) -- (#4,-#5);
}

\newcommand{\fclayer}[4]{%

	\foreach \k in {1,...,#1}
	{
		\node[draw,shape=circle,fill=Black!50,scale=0.05] (l#4n\k) at (0.75 * \k / #1 + #2,-0.75 * \k / #1 - #3) {};
	}
}

\newcommand{\chromosome}[3]
{%
	\subcaptionbox{#1}
	{
		\begin{minipage}{#2}
			\tiny
			#3
		\end{minipage}
	}
}

\newcommand{\slantedscope}[4]{%

	\pgfmathsetmacro\shiftx{#1 * 4}
	\pgfmathsetmacro\slantx{#2}
	\pgfmathsetmacro\shifty{#3 * 4}
	\pgfmathsetmacro\slanty{#4}

	\scoped[
		scale=0.2,
		yshift=\shifty cm,
		xshift=\shiftx cm,
		xslant=\slantx,
		yslant=\slanty
		]
}

\newcommand{\kernel}[7]{%

	\pgfmathsetmacro\kxbegin{int((#2 - #4) / 2)}
	\pgfmathsetmacro\kybegin{int((#3 - #5) / 2)}

	\pgfmathsetmacro\kxend{int(#2 - \kxbegin)}
	\pgfmathsetmacro\kyend{int(#2 - \kybegin)}

	\pgfmathsetmacro\centertilebegin{int(#2 / 2)}
	\pgfmathsetmacro\centertileend{\centertilebegin + 1}

	\slantedscope{0}{0.6}{#1}{0}{
		\fill[white,fill opacity=0.75] (0,0) rectangle (#2,#3);
		\draw[step=1,Black!15] (0,0) grid (#2,#3);
		\node[draw=none,rotate=60] at (#2 + 1,#3 - 2) {\tiny#7};
		\fill[#6,fill opacity=0.5] (\kxbegin,\kybegin) rectangle (\kxend,\kyend);
		\draw[black!40] (0,0) rectangle (#2,#2);
		\fill[#6,fill opacity=0.9] (\centertilebegin, \centertilebegin) rectangle (\centertileend, \centertileend);
	}
}

\newcommand{\convlayer}[6]
{%
	\foreach \k in {1,...,#1}
	{
		\filter{\k}{\k}{#1}{#2}{#3}{#4}{#5}{#6}
	}
}

\newcommand{\fclayerscope}[2]{%
	\scoped[
	shift={(#1,#2)},
	start chain = going right,
	node distance = 0pt
	]
}

\usepackage[american]{babel}
\usepackage[style=american]{csquotes}

\usepackage[%
style=ieee,
bibstyle=numeric,
useprefix=false,
maxnames=6,
uniquename=false,
firstinits,
isbn=false,
doi=false,
sorting=none,
backend=biber,
urldate=comp,
arxiv=abs,
alldates=year
]{biblatex}

\AtEveryBibitem{%
	\clearlist{address}
	\clearfield{date}
	\clearname{editor}
	\clearlist{publisher}
	\clearfield{isbn}
	\clearfield{issn}
	\clearfield{doi}
	\clearlist{location}
	\clearfield{month}
	\clearfield{series}
	\clearfield{primaryclass}
	\clearfield{eprintclass}
	\clearfield{volume}
	\clearfield{url}
}

\usepackage{xpatch}
\xpatchbibmacro{date+extrayear}{%
	\printtext[parens]%
}{%
	\setunit*{\addperiod\space}%
	\printtext%
}{}{}

\DeclareFieldFormat{titlecase}{\MakeSentenceCase{#1}}

\renewbibmacro{in:}{}

\newcommand{\red}[1]{\textcolor{Red}{#1}}

\usepackage[bottom]{footmisc}

\begin{document}

\title{Epigenetic evolution of deep convolutional models
\thanks{The first author is grateful for the Research Training Program (RTP) scholarship provided by the Australian Government.}
}

\author{\IEEEauthorblockN{Alexander Hadjiivanov}
\IEEEauthorblockA{\textit{School of Computer Science and Engineering} \\
University of New South Wales\\
Sydney, Australia \\
a.hadjiivanov@student.unsw.edu.au}
\and
\IEEEauthorblockN{Alan Blair}
\IEEEauthorblockA{\textit{School of Computer Science and Engineering} \\
University of New South Wales\\
Sydney, Australia \\
blair@cse.unsw.edu.au}

}

\maketitle

\begin{abstract}
In this study, we build upon a previously proposed neuroevolution framework to evolve deep convolutional models. Specifically, the genome encoding and the crossover operator are extended to make them applicable to layered networks. We also propose a convolutional layer layout which allows kernels of different shapes and sizes to coexist within the same layer, and present an argument as to why this may be beneficial. The proposed layout enables the size and shape of individual kernels within a convolutional layer to be evolved with a corresponding new mutation operator. The proposed framework employs a hybrid optimisation strategy involving structural changes through epigenetic evolution and weight update through backpropagation in a population-based setting. Experiments on several image classification benchmarks demonstrate that the crossover operator is sufficiently robust to produce increasingly performant offspring even when the parents are trained on only a small random subset of the training dataset in each epoch, thus providing direct confirmation that learned features and behaviour can be successfully transferred from parent networks to offspring in the next generation.
\end{abstract}

\section{Introduction}\label{sec:intro}
Neuroevolution (NE), or the process of evolving the structure and/or the weights of neural networks (NNs), has matured into a viable and versatile optimisation tool over the past three decades. Evolution tends to converge slowly and generally requires a large number of evaluations, so early work on NE was limited to relatively small networks \parencite{Wieland--1991,RichardsMoriartyMiikkulainen--1998}. As evolutionary algorithms grew in sophistication and the power and availability of hardware improved, NE was able to achieve excellent results in tasks of varying complexity \parencite{Yao--1999}, with a number of incremental improvements in genome encoding and algorithm efficiency (e.g., Symbiotic Adaptive NeuroEvolution (SANE) \parencite{RichardsMoriartyMiikkulainen--1998}, Enforced Sub-Populations (ESP) \parencite{GomezMiikkulainen--1999}, Evolution Strategy with Covariance Matrix Adaptation (CMA-ES) \parencite{HansenOstermeier--2001}, NeuroEvolution of Augmenting Topologies (NEAT) \parencite{StanleyMiikkulainen--2002} and Cooperative Synapse NeuroEvolution (CoSyNE) \parencite{GomezSchmidhuberMiikkulainen--2006}) seen around the turn of the century \parencite{FloreanoDuerrMattiussi--2008}. The success of NEAT gave rise to variants such as Hypercube-based NEAT (HyperNEAT), which uses NEAT to evolve a Compositional Pattern-Producing Network (CPPN) as a compact indirect encoding of the actual phenotype \parencite{StanleyDAmbrosioGauci--2009,SecretanBeatoDAmbrosioRodriguezCampbellEtAl--2011}. Interestingly, CPPNs can evolve not only the structure and weights but also the transfer function of the NN nodes, which is a somewhat rare mutation operation in NE. More recently, Cartesian Genetic Programming (CGP) has been applied to directed graphs (rather than tree-based structures, which are commonly used in GP) to evolve both feed-forward and recurrent NNs \parencite{KhanAhmadKhanMiller--2013} as well as heterogeneous networks with evolved transfer functions (similar to CPPNs) \parencite{TurnerMiller--2014}, achieving excellent results on dynamic control and classification tasks.

Within the last few years, research on NE has expanded from tasks which could be solved by evolved networks with relatively few weights, such as pole balancing \parencite{Igel--2003} and robot maze navigation in a synthetic environment \parencite{LehmanStanley--2011}, to more complex visual tasks such as image classification \parencite{VerbancsicsHarguess--2013}. Importantly, recognising the advantages of evolution as a global optimiser, there has been a paradigm shift towards utilising NE as an optimiser for the network structure in combination with backpropagation (BP) to fine-tune the network weights. For instance, deep convolutional NNs (CNNs) with multiple layers and millions of parameters have been evolved for tasks ranging from image classification \parencite{RealMooreSelleSaxenaSuematsuEtAl--2017,MiikkulainenLiangMeyersonRawalFinkEtAl--2017}, image captioning \parencite{MiikkulainenLiangMeyersonRawalFinkEtAl--2017} (using an evolved deep Long Short-Term Memory (LSTM) network) and even applications in particle physics (neutron scattering model selection) \parencite{YoungRoseJohnstonHellerKarnowskiEtAl--2017}. A differentiable version of CPPN was proposed in \parencite{FernandoBanarseReynoldsBessePfauEtAl--2016} to efficiently compress the representation of deep CNNs. Furthermore, genetic algorithms \parencite{XieYuille--2017}, particle swarm optimisation (PSO) \parencite{WangSunXueZhang--2018} and GP \parencite{SuganumaShirakawaNagao--2018} have also demonstrated excellent results on searching for optimal CNN structures for image classification tasks. In \parencite{NayebiBearKubiliusKarGanguliEtAl--2018}, an additional degree of complexity was explored by allowing local and long-range recurrent feedback connections to be discovered by evolution. As an extreme example of parallel NE, in \parencite{Desell--2017} a massively parallel distributed environment was set up on top of the BOINC\footnote{The Berkeley Open Infrastructure for Network Computing (\url{https://boinc.berkeley.edu}) is a generic distributed computing platform.} platform to evolve highly optimised convolutional networks that achieve state-of-the-art performance on the MNIST dataset with a reported accuracy of 99.43\%.

Evolution is commonly used to optimise the network architecture and minimise the number of parameters without compromising performance. For example, evolved CNN topologies have achieved the highest accuracy on an image classification task (Fashion-MNIST) compared to ten other popular models (including AlexNet and GoogLeNet) with a small fraction of the weights of the largest compared model (GoogLeNet) \parencite{SunXueZhang--2017}. Similar results (up to 12-fold reduction in the number of parameters without loss of accuracy) have been achieved through iterative connection pruning and retraining of large pretrained models (such as VGG-16 and AlexNet) on image classification tasks \parencite{HanPoolTranDally--2015}. Another interesting approach proposed recently is MetaQNN \parencite{BakerGuptaNaikRaskar--2017}, where state-of-the-art performance on image classification tasks has been achieved with simpler network architectures discovered through reinforcement learning (RL). Although not using evolution, the latter two examples clearly demonstrate the merit of architecture search and further strengthen the case for exploring new avenues for research on NE. Finally, despite the fact that evolution is usually used for optimising the structure rather than the weights, it has been recently demonstrated \parencite{SalimansHoChenSidorSutskever--2017} that evolution strategies can successfully optimise the weights of a NN with millions of parameters, rivalling the results obtained with RL on a number of OpenAI Gym 3D tasks and Atari games while offering advantages such as higher performance under distributed training and lower sensitivity to the temporal scale of the simulation.

\section{Background}\label{sec:background}
In previous research \parencite{HadjiivanovBlair--2016}, we proposed a NE framework (named Cortex) which is based on a direct genotype encoding using the ordered number of nodes in an unstructured network as a straightforward metric for matching network topologies during crossover. In this study, Cortex is extended to make it applicable to deep layered NNs, with particular focus on deep CNNs. This section provides a brief overview of a typical CNN architecture and the Cortex NE framework (particularly genome encoding, speciation and crossover), which are extended in several ways in \ref{sec:deep_NE} to make them applicable to deep CNN evolution.

\subsection{Convolutional networks}\label{sec:CNNs}
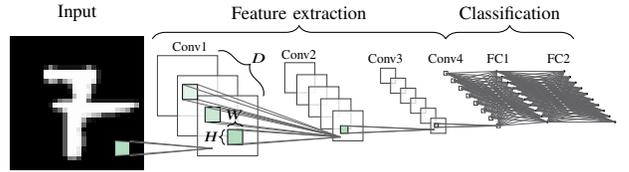
\begin{figure}
	\centering
	\begin{tikzpicture}

	\pgfmathsetmacro\xsize{8}
	\pgfmathsetmacro\ysize{8}

	\node at (0,-0.5) {\includegraphics[width=0.2\linewidth]{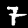}};
	\node at (0,0.7) {\scriptsize{Input}};

	\node at (1.5,0.25) {\tiny{Conv1}};
	\node at (2.95,0.15) {\tiny{Conv2}};
	\node at (4.1,0.1) {\tiny{Conv3}};
	\node at (4.9,0.1) {\tiny{Conv4}};
	\node at (5.6,0.1) {\tiny{FC1}};
	\node at (6.4,0.1) {\tiny{FC2}};

	\convlayer{3}{1.5}{0}{\xsize/2}{\ysize/2}{1}
	\convlayer{5}{3.5}{0}{\xsize/4}{\ysize/4}{2}
	\convlayer{6}{5}{0}{\xsize/8}{\ysize/8}{3}
	\convlayer{8}{6}{0}{\xsize/32}{\ysize/32}{4}
	\fclayer{10}{5.5}{0}{5}
	\fclayer{10}{6.4}{0}{6}

	\patch{0.5}{1}{0.2}{1.8}{1.1}{Green!30}

	\foreach \p in {0,...,2}
	\pgfmathsetmacro\opacity{30 - \p * 10}
	\patch{2 - 0.15 * \p * \xsize/4}{0.85 + 0.15 * \p * \ysize/4}{0.2}{3.5}{0.95}{Green!\opacity};

	\patch{3.5}{0.8}{0.1}{4.75}{0.85}{Green!30}
	\patch{4.75}{0.77}{0.05}{5.6}{0.8}{Green!30}

	\foreach \m in {1,...,8}
	\foreach \n in {1,...,10}
	\draw[ultra thin,Black!75] (l4n\m) -- (l5n\n);

	\foreach \m in {1,...,10}
	\foreach \n in {1,...,10}
	\draw[ultra thin,Black!75] (l5n\m) -- (l6n\n);

	\draw[decorate,decoration={brace,amplitude=5pt,raise=3pt}]
	(1,0.25) to node[black,midway,above=7pt] {\scriptsize{Feature extraction}} (4.9,0.25);

	\draw[decorate,decoration={brace,amplitude=5pt,raise=3pt}]
	(4.9,0.25) to node[black,midway,above=7pt] {\scriptsize{Classification}} (6.6,0.25);

	\draw[decorate,decoration={brace,amplitude=5pt,raise=3pt}]
	(1.8,0.12) to node[black,midway,xshift=9pt,yshift=8pt] {\tiny{$D$}} (2.4,-0.48);

	\draw[decorate,decoration={brace,amplitude=2pt,raise=1pt}]
	(2,-0.85) to node[black,midway,above=0pt] {\tiny{$W$}} (2.2,-0.85);

	\draw[decorate,decoration={brace,amplitude=2pt,raise=1pt}]
	(2,-1.05) to node[black,midway,xshift=-6pt] {\tiny{$H$}} (2,-0.85);
	\end{tikzpicture}

	\caption[]{A typical convolutional NN architecture. The first layer convolves the input with a set of kernels (filters) to obtain a shift-invariant response map of the input. The kernels in each convolutional layer have dimensionality $D\times W\times H$, where $W$ and $H$ are the width and height of the kernel and $D$ is the number of input channels (i.e., the convolution `depth'). By convention, all kernels in a convolutional layer have the same dimensions, so the kernel size effectively becomes an attribute of the layer itself. In addition, the number of channels is taken as the number of input maps in the previous layer. In the example above, $D$ represents the number of channels for kernels in layer $Conv2$. The classification part consists of one or more FC layers.}\label{fig:typical_conv_net}
\end{figure}

Convolutional networks have enjoyed an exponential surge in popularity since demonstrating extremely high performance in various domains of machine learning (particularly classification tasks relying on pattern recognition), such as document recognition \parencite{LecunBottouBengioHaffner--1998} and image classification \parencite{KrizhevskySutskeverHinton--2012,SzegedyLiuJiaSermanetReedEtAl--2015}. The canonical convolutional network architecture consists of one or more feature extraction layers followed by one or more classification (fully connected; FC) layers (Fig.~\ref{fig:typical_conv_net}).

\begin{figure}
	\subcaptionbox{\label{subfig:conv.k3s1p0}}[0.3\columnwidth]
	{
		\centering
		\begin{tikzpicture}
		\fclayerscope{0.6em}{1}
		{
			\foreach \m in {1,...,5}
			\node[LayerNode] (node\m) {};

			\node[black,above of=node3,yshift=10pt] {\footnotesize{$K:3,S:1,P:0$}};
		}
		\fclayerscope{0}{0}
		{
			\foreach \m in {1,2,3}
			\node[LayerNode,fill=Green!40] (input\m) {};
			\foreach \m in {4,...,7}
			\node[LayerNode,fill=Black!20] (input\m) {};
		}

		\foreach \m in {1,...,5}
		{
			\foreach \n in {0,...,2}
			{
				\pgfmathsetmacro\sumtwo{int(\m + \n)}
				\draw (node\m.south) -- (input\sumtwo.north);
			}
		}

		\draw[decorate,decoration={brace, amplitude=5pt,raise=1pt,mirror}]
		(input1.south west) to node[black,midway,below= 5pt] {\footnotesize{Input}} (input7.south east);

		\node[White, below of=input5,yshift=-1em] {};

		\end{tikzpicture}
	}
	\subcaptionbox{\label{subfig:conv.k3s2p0}}[0.3\columnwidth]
	{
		\centering
		\begin{tikzpicture}
		\fclayerscope{1.2em}{1}
		{
			\foreach \m in {1,...,3}
			\node[LayerNode] (node\m) {};

			\node[black,above of=node2,yshift=10pt] {\footnotesize{$K:3,S:2,P:0$}};
		}
		\fclayerscope{0}{0}
		{
			\foreach \m in {1,2,3}
			\node[LayerNode,fill=Green!40] (input\m) {};
			\foreach \m in {4,...,7}
			\node[LayerNode,fill=Black!20] (input\m) {};
		}

		\foreach \m in {1,2,3}
		{
			\foreach \n in {0,...,2}
			{
				\pgfmathsetmacro\sumtwo{int(2 * \m + \n - 1)}
				\draw (node\m.south) -- (input\sumtwo.north);
			}
		}

		\draw[decorate,decoration={brace, amplitude=5pt,raise=1pt,mirror}]
		(input1.south west) to node[black,midway,below= 5pt] {\footnotesize{Input}} (input7.south east);

		\node[White, below of=input5,yshift=-1em] {};

		\end{tikzpicture}
	}
	\subcaptionbox{\label{subfig:conv.k3s1p1}}[0.3\columnwidth]
	{
		\centering
		\begin{tikzpicture}
		\fclayerscope{0.6em}{1}
		{
			\foreach \m in {1,...,7}
			\node[LayerNode] (node\m) {};

			\node[black,above of=node4,yshift=10pt] {\footnotesize{$K:3,S:1,P:1$}};
		}
		\fclayerscope{0}{0}
		{
			\node[LayerNode,fill=Green!20] (input1) {};
			\foreach \m in {2,3}
			\node[LayerNode,fill=Green!40] (input\m) {};
			\foreach \m in {4,...,8}
			\node[LayerNode,fill=Black!20] (input\m) {};
			\node[LayerNode] (input9) {};
		}

		\foreach \m in {1,...,7}
		{
			\foreach \n in {0,...,2}
			{
				\pgfmathsetmacro\sumtwo{int(\m + \n)}
				\draw (node\m.south) -- (input\sumtwo.north);
			}
		}

		\draw[decorate,decoration={brace, amplitude=5pt,raise=1pt,mirror}]
		(input2.south west) to node[black,midway,below= 5pt] {\footnotesize{Input}} (input8.south east);

		\node[Black, below of=input5,yshift=-0.75em] (zp) {\footnotesize{Zero padding}};
		\draw[->] (zp) -- (input1.south);
		\draw[->] (zp) -- (input9.south);

		\end{tikzpicture}
	}
	\caption[]{An illustration of a convolution operation in one dimension. A kernel with size $K$ is slid across the input, and the dot product of the kernel and the patch of the input that it covers is computed as the response of the kernel for that patch. The stride $S$ determines how many elements the kernel is shifted by at each step, and the padding $P$ determines the offset of the convolution operation (i.e., whether or not it is aligned with the edge of the input). Both $S$ and $P$ determine the size of the response. (\subref{subfig:conv.k3s1p0}\textasciitilde\subref{subfig:conv.k3s1p1}) Kernel responses for different values of $K$, $S$ and $P$.}\label{fig:ksp}
\end{figure}
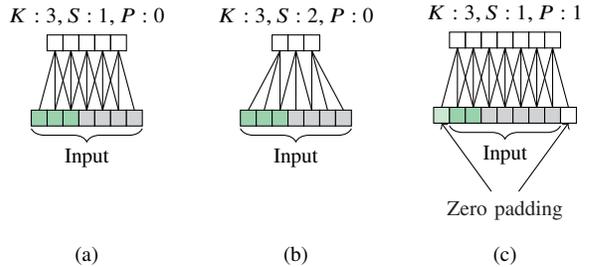

Convolutional layers are characterised by their kernel size $K$, stride $S$ and padding $P$ (Fig. \ref{fig:ksp}), where the kernel size is regarded as a property of the layer itself. In this study, the extent of the kernel in the width and height dimensions is taken as a property of each individual kernel, which enables kernels of different sizes and shapes to be evolved.

\subsection{Cortex NE framework}\label{sec:cortex}
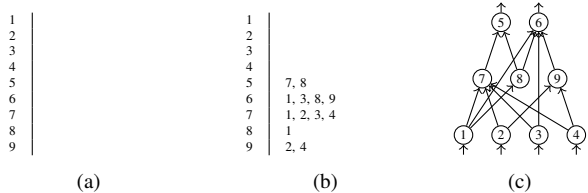
\begin{figure}
	\centering
	\chromosome{\label{subfig:cortex.genome}}{0.3\columnwidth}
	{
		\begin{tabular}{r|l}
			1 &   \\
			2 &   \\
			3 &   \\
			4 &   \\
			5 &   \\
			6 &   \\
			7 &   \\
			8 &   \\
			9 &
		\end{tabular}
	}
	\chromosome{\label{subfig:cortex.genotype}}{0.3\columnwidth}
	{
		\begin{tabular}{r|l}
			1 &            \\
			2 &            \\
			3 &            \\
			4 &            \\
			5 & 7, 8       \\
			6 & 1, 3, 8, 9 \\
			7 & 1, 2, 3, 4 \\
			8 & 1          \\
			9 & 2, 4
		\end{tabular}
	}
	\subcaptionbox{\label{subfig:cortex.phenotype}}[0.2\columnwidth]
	{
		\centering
		\begin{tikzpicture}[xscale=0.5,yscale=0.75]

		\foreach \l [count=\i] in {1,2,3,4}
		\node [Node] (i-\i) at (\i - 1,0) {\l};

		\foreach \l [count=\i] in {7,8,9}
		\node [Node] (h-\i) at (\i - 0.5,1) {\l};

		\foreach \l [count=\i] in {5,6}
		\node [Node] (o-\i) at (\i,2) {\l};

		\foreach \l [count=\i] in {1,...,4}
		\draw [<-] (i-\i.south) -- ([yshift=-0.5em] i-\i.south) {};

		\foreach \l [count=\i] in {1,2}
		\draw [->] (o-\i.north) -- ([yshift=0.5em] o-\i.north) {};

		\draw[Link] (i-1) to (h-1);
		\draw[Link] (i-1) to (h-2);
		\draw[Link] (i-1) to (o-2);
		\draw[Link] (i-2) to (h-1);
		\draw[Link] (i-2) to (h-3);
		\draw[Link] (i-3) to (h-1);
		\draw[Link] (i-3) to (o-2);
		\draw[Link] (i-4) to (h-1);
		\draw[Link] (i-4) to (h-3);
		\draw[Link] (h-1) to (o-1);
		\draw[Link] (h-2) to (o-1);
		\draw[Link] (h-2) to (o-2);
		\draw[Link] (h-3) to (o-2);

		\end{tikzpicture}
	}
	\caption[]{(\subref{subfig:cortex.genome}) A Cortex genome (an ordered array of nodes), (\subref{subfig:cortex.genotype}) a corresponding genotype (a genome with added inward connections, but no weights assigned to them) and (\subref{subfig:cortex.phenotype}) a phenotype with weights assigned to all connections.}
	\label{fig:cortex.taxonomy}
\end{figure}

In Cortex, a distinction is made between a genome (a `skeleton' of ordered nodes shared by all networks with the same number of nodes; Fig. \ref{subfig:cortex.genome}), a genotype (a genome with added connections but without weight information; Fig. \ref{subfig:cortex.genotype}) and a phenotype (a genotype with weights assigned to all connections; Fig. \ref{subfig:cortex.phenotype}). The genome is used for \textit{speciation}, which is a form of niching introduced together with NEAT \parencite{Stanley--2004} as way to group similar network genomes. Speciation was designed to improve the chance of survival of networks after structural mutations (such as adding or removing nodes or connections), which are likely to reduce the network's fitness initially, even though they might be beneficial in the long run. Using this type of speciation, networks compete only within the same species rather than the entire ecosystem\footnote{In Cortex, a group of networks belonging to the same species is referred to as a \textit{population}, and a group of two or more coexisting populations is referred to as an \textit{ecosystem}.}, which increases the chance of survival of unfit networks.

The genotypes defined by a particular genome (species) are used for crossover, which is convenient because the nodes in the genotype are \textit{ordered}, meaning that all genotypes can be aligned automatically. Although the only difference between a genotype and a phenotype is that in the latter all connections have weights assigned to them, the weights do not play a role in determining which genes are transferred to the offspring during crossover. Connections are inherited by sampling the parent genotypes with probabilities proportional to the parents' relative fitness values (cf. \eqref{eq:relative_fitness}). If a connection exists in both parents, the offspring is more likely to inherit the connection weight from the fitter parent, whereas if a connection exists in only one parent, that parent's relative fitness is used as a probability to check if the connection should be inherited or skipped altogether.

In Cortex, the number of nodes in the genome is used as the sole criterion for speciation. A welcome side effect of this speciation technique is that the ecosystem can be initialised with more than one species from the onset, which facilitates exploration and promotes diversity.

\begin{figure}
	\centering
	\subcaptionbox{Parent 1\label{subfig:cortex.crossover.parent1}}[0.3\columnwidth]
	{
		\centering
		\begin{tikzpicture}[baseline,xscale=0.5,yscale=0.75]

		\foreach \l [count=\i] in {1,2,3,4}
		\node [Node] (bi-\i) at (\i,0) {\l};

		\foreach \l [count=\i] in {7,8,9}
		\node [Node] (bh-\i) at (\i + 0.5,1) {\l};

		\foreach \l [count=\i] in {5,6}
		\node [Node] (bo-\i) at (\i + 1,2) {\l};

		\foreach \l [count=\i] in {1,...,4}
		\draw [<-] (bi-\i.south) -- ([yshift=-0.5em] bi-\i.south) {};

		\foreach \l [count=\i] in {1,2}
		\draw [->] (bo-\i.north) -- ([yshift=0.5em] bo-\i.north) {};

		\draw[Link] (bi-1) to (bh-1);
		\draw[Link] (bi-1) to (bh-2);
		\draw[Link] (bi-1) to (bo-2);
		\draw[Link] (bi-2) to (bh-1);
		\draw[Link] (bi-2) to (bh-3);
		\draw[Link] (bi-3) to (bh-1);
		\draw[Link] (bi-3) to (bo-2);
		\draw[Link] (bi-4) to (bh-1);
		\draw[Link] (bi-4) to (bh-3);
		\draw[Link] (bh-1) to (bo-1);
		\draw[Link] (bh-2) to (bo-1);
		\draw[Link] (bh-2) to (bo-2);
		\draw[Link] (bh-3) to (bo-2);
		\end{tikzpicture}

		\vspace{0.5em}
		{
			\tiny
			\centering
			\begin{tabular}[b]{r|l}
				1 &            \\
				2 &            \\
				3 &            \\
				4 &            \\
				5 & 7, 8       \\
				6 & 1, 3, 8, 9 \\
				7 & 1, 2, 3, 4 \\
				8 & 1          \\
				9 & 2, 4
			\end{tabular}
		}
	}
	\subcaptionbox{Parent 2\label{subfig:cortex.crossover.parent2}}[0.3\columnwidth]
	{
		\centering
		\begin{tikzpicture}[xscale=0.5,yscale=0.75]

		\foreach \l [count=\i] in {1,2,3,4}
		\node [Node] (ri-\i) at (\i,0) {\l};

		\foreach \l [count=\i] in {7,8,9}
		\node [Node] (rh-\i) at (\i + 0.5,1) {\l};

		\foreach \l [count=\i] in {5,6}
		\node [Node] (ro-\i) at (\i + 1,2) {\l};

		\foreach \l [count=\i] in {1,...,4}
		\draw [<-] (ri-\i.south) -- ([yshift=-0.5em] ri-\i.south) {};

		\foreach \l [count=\i] in {1,2}
		\draw [->] (ro-\i.north) -- ([yshift=0.5em] ro-\i.north) {};

		\draw[Link,Red] (ri-1) to (rh-1);
		\draw[Link,Red] (ri-1) to (rh-3);
		\draw[Link,Red] (ri-2) to (rh-2);
		\draw[Link,Red] (ri-2) to (rh-3);
		\draw[Link,Red] (ri-3) to (rh-1);
		\draw[Link,Red] (ri-3) to (rh-2);
		\draw[Link,Red] (ri-4) to (rh-1);
		\draw[Link,Red] (ri-4) to (rh-3);
		\draw[Link,Red] (rh-1) to (ro-1);
		\draw[Link,Red] (rh-2) to (ro-2);
		\draw[Link,Red] (rh-3) to (ro-2);

		\end{tikzpicture}

		\vspace{0.5em}
		{
			\tiny
			\centering
			\begin{tabular}[b]{r|l}
				1 &                           \\
				2 &                           \\
				3 &                           \\
				4 &                           \\
				5 & \red{7}                   \\
				6 & \red{8}, \red{9}          \\
				7 & \red{1}, \red{3}, \red{4} \\
				8 & \red{2}, \red{3}          \\
				9 & \red{1}, \red{2}, \red{4}
			\end{tabular}
		}
	}
	\subcaptionbox{Offspring\label{subfig:cortex.crossover.offspring}}[0.3\columnwidth]
	{
		\centering
		\begin{tikzpicture}[xscale=0.5,yscale=0.75]

		\foreach \l [count=\i] in {1,2,3,4}
		\node [Node] (off_i-\i) at (\i,0) {\l};

		\foreach \l [count=\i] in {7,8,9}
		\node [Node] (off_h-\i) at (\i + 0.5,1) {\l};

		\foreach \l [count=\i] in {5,6}
		\node [Node] (off_o-\i) at (\i + 1,2) {\l};

		\foreach \l [count=\i] in {1,...,4}
		\draw [<-] (off_i-\i.south) -- ([yshift=-0.5em] off_i-\i.south) {};

		\foreach \l [count=\i] in {1,2}
		\draw [->] (off_o-\i.north) -- ([yshift=0.5em] off_o-\i.north) {};

		\draw[Link] (off_i-1) to (off_o-2);
		\draw[Link] (off_i-1) to (off_h-1);
		\draw[Link] (off_i-1) to (off_h-2);
		\draw[Link,Red] (off_i-2) to (off_h-2);
		\draw[Link,Red] (off_i-2) to (off_h-3);
		\draw[Link] (off_i-3) to (off_o-2);
		\draw[Link] (off_i-3) to (off_h-1);
		\draw[Link,Red] (off_i-4) to (off_h-1);
		\draw[Link] (off_i-4) to (off_h-3);
		\draw[Link] (off_h-1) to (off_o-1);
		\draw[Link] (off_h-2) to (off_o-1);
		\draw[Link,Red] (off_h-2) to (off_o-2);
		\draw[Link,Red] (off_h-3) to (off_o-2);
		\end{tikzpicture}

		\vspace{0.5em}
		{
			\tiny
			\centering
			\begin{tabular}[b]{r|l}
				1 &                        \\
				2 &                        \\
				3 &                        \\
				4 &                        \\
				5 & 7, 8                   \\
				6 & 1, 3, \red{8}, \red{9} \\
				7 & 1, 3, \red{4}          \\
				8 & 1, \red{2}             \\
				9 & \red{2}, 4
			\end{tabular}
		}
	}
	\caption[]{Crossover operation between unstructured parent networks with a matching number of nodes. (\subref{subfig:cortex.crossover.parent1}, \subref{subfig:cortex.crossover.parent2}) Parents participating in the crossover and (\subref{subfig:cortex.crossover.offspring}) the resulting offspring. Nodes in the genotype are ordered, which aids with topology matching during crossover.}
	\label{fig:cortex_crossover}
\end{figure}
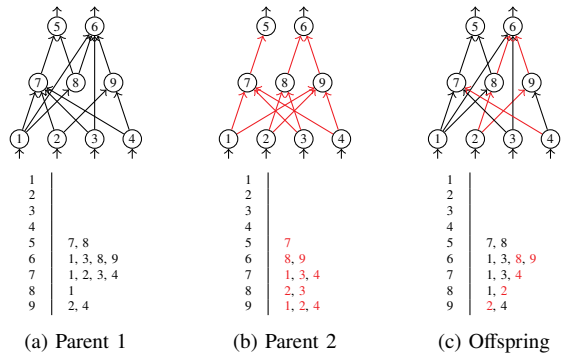

Arguably, the most beneficial aspect of matching network topologies based on the ordered number of nodes is in regard to crossover, which is guaranteed to produce functional offspring containing \textit{only} genes from the parents, without the need to introduce new connections or prune existing ones (Fig.~\ref{fig:cortex_crossover}). Although the benefits of this are less obvious for unstructured networks since in general they do not impose any restrictions on the dimensionality of the fan-in and the fan-out of individual nodes, it becomes important for layered networks, where every layer expects the input to have a certain shape and size. This issue was considered, for example, in \parencite{YoungRoseJohnstonHellerKarnowskiEtAl--2017}, where the fan-in dimensionality of layer modules of incompatible size was adjusted in order to produce a functional model.

It should be noted that in Cortex weights are inherited unaltered from the parent networks rather than being initialised at every generation. In this context, the evolution mode in Cortex is epigenetic (or Lamarckian) since offspring effectively inherit traits that the parent networks have acquired during their lifetime.


\section{Evolving deep convolutional models}\label{sec:deep_NE}
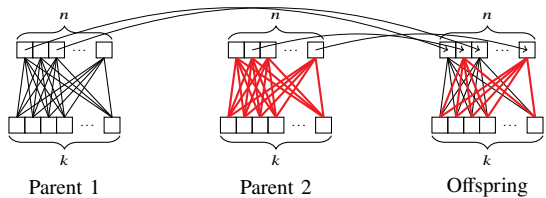
\begin{figure}
	\centering

	\begin{tikzpicture}

	\fclayerscope{0.3em}{1}{
		\foreach \m in {1,...,5}
		\ifthenelse{\equal{\m}{4}}
		{\node[MissingNode] (p1node\m) {\tiny{\ldots}};}
		{\node[LayerNode] (p1node\m) {};};

		\draw[decorate,decoration={brace, amplitude=5pt,raise=1pt}]
		(p1node1.north west) to node[black,midway,above= 5pt] {\tiny{$n$}} (p1node5.north east);
	}

	\fclayerscope{0em}{0}{
		\foreach \m in {1,...,6}
		\ifthenelse{\equal{\m}{5}}
		{\node[MissingNode] (p1input\m) {\tiny{\ldots}};}
		{\node[LayerNode] (p1input\m) {};};

		\foreach \m in {1,...,5}
		\foreach \n in {1,...,4,6}
		\ifthenelse{\equal{\m}{4} \OR \equal{\n}{5}}
		{}
		{\draw (p1node\m.south) -- (p1input\n.north);};

		\draw[decorate,decoration={brace, amplitude=5pt,raise=1pt,mirror}]
		(p1input1.south west) to node[black,midway,below= 5pt] (p1k) {\tiny{$k$}} (p1input6.south east);

		\node[black,below of=p1k,yshift=-1em] {\footnotesize{Parent 1}};
	}

	\fclayerscope{8.3em}{1}{
		\foreach \m in {1,...,5}
		\ifthenelse{\equal{\m}{4}}
		{\node[MissingNode] (p2node\m) {\tiny{\ldots}};}
		{\node[LayerNode] (p2node\m) {};};

		\draw[decorate,decoration={brace, amplitude=5pt,raise=1pt}]
		(p2node1.north west) to node[black,midway,above= 5pt] {\tiny{$n$}} (p2node5.north east);
	}

	\fclayerscope{8em}{0}{
		\foreach \m in {1,...,6}
		\ifthenelse{\equal{\m}{5}}
		{\node[MissingNode] (p2input\m) {\tiny{\ldots}};}
		{\node[LayerNode] (p2input\m) {};};

		\foreach \m in {1,...,5}
		\foreach \n in {1,...,4,6}
		\ifthenelse{\equal{\m}{4} \OR \equal{\n}{5}}
		{}
		{\draw[Red,thick] (p2node\m.south) -- (p2input\n.north);};

		\draw[decorate,decoration={brace,amplitude=5pt,raise=1pt,mirror}]
		(p2input1.south west) to node[black,midway,below= 5pt] (p2k) {\tiny{$k$}} (p2input6.south east);

		\node[black,below of=p2k,yshift=-1em] {\footnotesize{Parent 2}};
	}

	\fclayerscope{16.3em}{1}
	{
		\foreach \m in {1,...,5}
		\ifthenelse{\equal{\m}{4}}
		{\node[MissingNode] (onode\m) {\tiny{\ldots}};}
		{\node[LayerNode] (onode\m) {};};

		\draw[decorate,decoration={brace, amplitude=5pt,raise=1pt}]
		(onode1.north west) to node[black,midway,above= 5pt] {\tiny{$n$}} (onode5.north east);
	}

	\fclayerscope{16em}{0}{
		\foreach \m in {1,...,6}
		\ifthenelse{\equal{\m}{5}}
		{\node[MissingNode] (oinput\m) {\tiny{\ldots}};}
		{\node[LayerNode] (oinput\m) {};};

		\foreach \m in {1,3}
		\foreach \n in {1,...,4,6}
		\draw (onode\m.south) -- (oinput\n.north);

		\foreach \m in {2,5}
		\foreach \n in {1,...,4,6}
		\draw[Red,thick] (onode\m.south) -- (oinput\n.north);

		\draw[decorate,decoration={brace, amplitude=5pt,raise=1pt,mirror}]
		(oinput1.south west) to node[black,midway,below= 5pt] (ok) {\tiny{$k$}} (oinput6.south east);

		\node[black,below of=ok,yshift=-1em] {\footnotesize{Offspring}};
	}

	\draw[->,ultra thin,bend left=20] (p1node1.center) to node[draw=none,midway,above] {} (onode1.center);
	\draw[->,ultra thin,bend left=20] (p1node3.center) to node[draw=none,midway,above] {} (onode3.center);

	\draw[->,ultra thin,bend left=15] (p2node2.center) to node[draw=none,midway,above] {} (onode2.center);
	\draw[->,ultra thin,bend left=15] (p2node5.center) to node[draw=none,midway,above] {} (onode5.center);

	\end{tikzpicture}

	\caption[]{Crossover between layered networks with matching number of nodes in each layer is guaranteed to preserve the dimensionality of each layer's input. This means that nodes can be transferred together with all of their input connections from the parents to the offspring without modification. In the case of convolutional layers, each kernel represents a node that can be inherited from the parent. Therefore, being able to manipulate individual kernels in a convolutional layer is essential (cf. \ref{sec:deep_NE.mutation.conv_layers}).}
	\label{fig:layered_model_crossover}
\end{figure}

In this study, the above topology matching and speciation scheme is extended to layered networks by treating layers as \textit{chromosomes} containing a number of genes (nodes). Two species are considered identical if they have the same number of layers of each type and each layer contains the same number of nodes. Furthermore, during crossover nodes together with \textit{all of their input connections} are treated as indelible genes which can be transferred unaltered from the parents to the offspring (Fig.~\ref{fig:layered_model_crossover}). If speciation is used, the input dimensionality of nodes and layers is preserved, and there is no need to add or remove any nodes or input connections in the offspring. However, nodes and/or input connections may need to be added to or removed from the offspring if speciation is not used, which may negatively impact the fitness of new offspring. To test this intuition, the experiments presented in \ref{sec:experiments} were performed with speciation enabled and disabled.

In the proposed framework, a population of deep CNNs is initialised and subsequently evolved over a number of generations. A single generation consists of several procedures, starting with training and evaluating all networks, followed by calibration, evolution (mutation and crossover), and finally culling. These procedures are described in more detail below.

\subsection{Ecosystem initialisation}\label{sec:deep_NE.initialisation}
With speciation enabled, the initial networks in the ecosystem are distributed evenly among the initial number of species, where each network in a species is initialised according to the species genome. The first species always has a minimal genome (containing only an output layer looking directly at the input). Each subsequent species is generated from an isolated network (i.e., a randomly generated network which does not belong to any species) which is mutated randomly until its genome does not match that of any of the existing species. The isolated network's genome is then used to generate a new species and a corresponding population of networks, and so on until the preset number of initial species is reached.

With speciation disabled, the initial ecosystem is essentially treated as a single species regardless of the genome. Networks are generated one at a time with a minimal genome (just an output layer), and a random mutation is applied to each network to promote initial diversity.

The shape of each kernel is initialised by sampling a weighted distribution of kernel shapes. Each shape is assigned a probability weighting $p_{w,h}$ inversely proportional to the area of the corresponding kernel computed as the product of its width and height dimensions $d_{w}$ and $d_{h}$ (not considering the number of channels):
\begin{equation}
	p_{w,h}=\exp(-\prod_{w,h}{d_{w}d_{h}})\label{eq:kernel_stride_init_size}
\end{equation}
For example, a kernel of size $3\times3$ would have a probability weighting of $\exp(-9)$ ($\approx0.00012$). The same procedure is used to initialise the stride of convolutional layers. The motivation for this initialisation method is to ensure that most networks start with minimal kernels and strides and increase them over the course of the evolution. Kernel and stride mutations are described in more detail in \ref{sec:deep_NE.mutation.conv_layers}.

\subsection{Training and evaluation}\label{sec:deep_NE.training}
All networks are trained with standard BP with Adadelta as the optimiser by default, which is chosen because it does not require the learning rate to be set manually. The classification accuracy on the test set is used as the absolute fitness.

\subsection{Calibration}\label{sec:deep_NE.calibration}
Before the evolution step, the fitness values and complexity of all networks are scaled to fall between $0$ and $1$ by computing the mean $\mu_{f}$ and standard deviation $\sigma_{f}$ of the fitness values of networks in the species (or the entire ecosystem if speciation is disabled). Then, the relative fitness $f_{r}(n_{i})$ for network $n_{i}$ is computed from its absolute fitness $f_{a}(n_{i})$ as follows:

\begin{equation}
f_{r}(n_{i}) = g\left(\frac{f_{a}(n_{i}) - \mu_{f}}{\sigma_{f}}\right)\label{eq:relative_fitness}
\end{equation}

where $g$ denotes the logistic function. Scaling the fitness to a value between $0$ and $1$ enables the relative fitness to be used as a probability for various random operations.

\subsection{Mutation}\label{sec:deep_NE.mutation}
Direct addition and removal of layers (both convolutional and FC) and nodes (in convolutional layers, a node is defined as a single kernel) is allowed, as well as changing the size of individual kernels and the stride of convolutional layers. For the kernel size and stride parameters, mutation is applied to a randomly selected dimension with probability as outlined below.

Structural mutations (adding or removing a node or a layer, resizing a kernel or resizing the stride parameter of a convolutional layer) are performed by sampling mutation types from a weighted probability distribution. The probability weighting of each mutation type is inversely proportional to an estimate of how many connections in the network the mutation would affect. For stride resizing, we use an estimate of how many \textit{output} nodes in convolutional layers would be affected:

\begin{equation}
	p_{layer} \propto \frac{1}{\mu_{c}(\mu_{n} + \sigma_{n})} \label{eq:layer_mutation_prob}
\end{equation}
\begin{equation}
	p_{node} \propto \frac{1}{\mu_{c}} \label{eq:node_mutation_prob}
\end{equation}
\begin{equation}
	p_{stride} \propto \frac{1}{N_{cl}\mu_{o}} \label{eq:stride_mutation_prob}
\end{equation}
\begin{equation}
	p_{kernel} \propto \frac{1}{N_{k}\mean{A_{k}}} \label{eq:kernel_mutation_prob}
\end{equation}

where $\mu_{c}$ is the mean number of connections per node, $\mu_{n}$ and $\sigma_{n}$ are the mean and standard deviation of the node count per layer, $N_{cl}$ is the number of convolutional layers, $\mu_{o}$ is the mean number of output nodes per convolutional layer, $N_{k}$ is the mean number of kernels\footnote{The experiments below were conducted with the mean number of all nodes ($N_{n}$) instead of $N_{k}$ due to an oversight. However, in practical terms this did not significantly change the ratio of the corresponding weights in the weighted distribution.}, and $\mean{A_{k}}$ is the mean kernel area ($W\times H$) in the network's convolutional layers. All of these probability weightings are computed by using statistics only for the network being mutated.

\subsection{Evolving convolutional layer parameters}\label{sec:deep_NE.mutation.conv_layers}
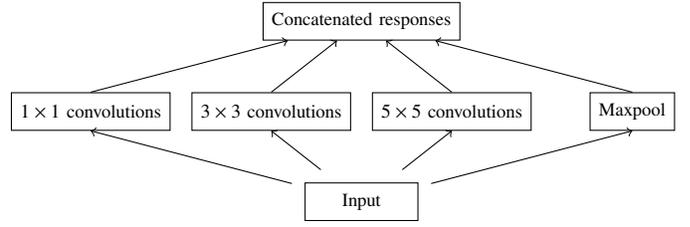
\begin{figure}
	\centering
	\begin{tikzpicture}[scale=0.6]
	\node (r1) at (6,2) [draw,minimum height=0.5cm] {\scriptsize{Concatenated responses}};
	\node (r2) at (0,0) [draw,minimum height=0.5cm] {\scriptsize{$1\times1$ convolutions}};
	\node (r3) at (4,0) [draw,minimum height=0.5cm] {\scriptsize{$3\times3$ convolutions}};
	\node (r4) at (8,0) [draw,minimum height=0.5cm] {\scriptsize{$5\times5$ convolutions}};
	\node (r5) at (12,0) [draw,minimum height=0.5cm] {\scriptsize{Maxpool}};
	\node (r6) at (6,-2) [draw,minimum width=1.5cm,minimum height=0.5cm,outer sep=5pt] {\scriptsize{Input}};

	\draw [->] (r6) -- (r2.south);
	\draw [->] (r6) -- (r3.south);
	\draw [->] (r6) -- (r4.south);
	\draw [->] (r6) -- (r5.south);

	\draw [->] (r2.north) -- (r1);
	\draw [->] (r3.north) -- (r1);
	\draw [->] (r4.north) -- (r1);
	\draw [->] (r5.north) -- (r1);
	\end{tikzpicture}
	\caption[]{A high-level representation of an Inception module. The input is convolved with kernels of multiple sizes, and the results are concatenated before being fed into the next layer.}
	\label{fig:inception_module}
\end{figure}

As mentioned above, it is generally assumed that the kernel size is a property of the layer rather than individual kernels. In other words, all kernels in the same layer usually have the same shape and size. However, the recently proposed Inception module architecture (Fig.~\ref{fig:inception_module}) \parencite{SzegedyLiuJiaSermanetReedEtAl--2015} breaks this trend by using kernels of different size ($1\times1$, $3\times3$ and $5\times5$ in the original paper) to convolve the input of some of the layers, and the outputs of all the kernels are concatenated to produce the final output of the layer. The GoogLeNet implementation of the Inception architecture won the 2014 ILSVRC\footnote{ImageNet Large Scale Visual Recognition Competition} challenge with a top-5 error rate of 5\%.

\begin{figure}
	\centering

	\subcaptionbox{\label{subfig:same_shape}}[0.45\columnwidth]
	{
		\begin{tikzpicture}

		\pgfmathsetmacro\xsize{7}
		\pgfmathsetmacro\ysize{7}

		\kernel{0}{\xsize}{\ysize}{7}{7}{Magenta}{$7\times7$}
		\kernel{2}{\xsize}{\ysize}{7}{7}{Blue}{$7\times7$}
		\kernel{3}{\xsize}{\ysize}{7}{7}{Green}{$7\times7$}
		\kernel{4}{\xsize}{\ysize}{7}{7}{Red}{$7\times7$}

		\node[draw=none,rotate=90] at (0.25,1) {\huge\ldots};
		\end{tikzpicture}
	}
	\subcaptionbox{\label{subfig:irregular_shape}}[0.45\columnwidth]
	{
		\begin{tikzpicture}

		\pgfmathsetmacro\xsize{7}
		\pgfmathsetmacro\ysize{7}

		\kernel{0}{\xsize}{\ysize}{5}{3}{Magenta}{$5\times3$}
		\kernel{2}{\xsize}{\ysize}{1}{7}{Blue}{$1\times7$}
		\kernel{3}{\xsize}{\ysize}{3}{5}{Green}{$3\times5$}
		\kernel{4}{\xsize}{\ysize}{3}{3}{Red}{$3\times3$}

		\node[draw=none,rotate=90] at (0.25,1) {\huge\ldots};

		\end{tikzpicture}
	}
	\caption[]{(\subref{subfig:same_shape}) A layer containing square kernels with the same size ($7\times7$ elements). The kernel size is essentially a property of the layer. (\subref{subfig:irregular_shape}) A layer with kernels of various shapes and sizes stacked along a line passing through the central element (marked as a darker element).}
	\label{fig:irregular_kernels}
\end{figure}
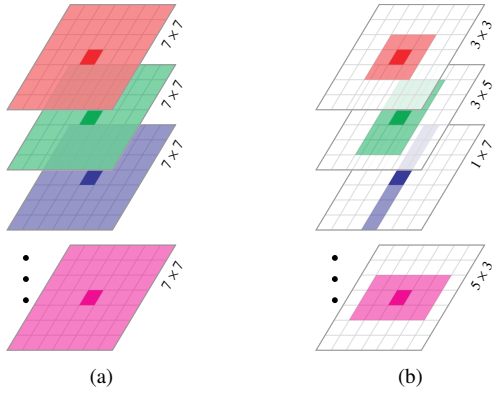

Inspired by the concept of convolving the input with kernels of various sizes, we go one step further and consider convolutional layers containing kernels of different \textit{shapes}, where the shape is taken as a property of individual kernels rather than the entire layer. Below, we use kernel \textit{shape} instead of kernel \textit{size} to indicate that the kernel might not be square. Specifically, under the mild assumption that kernels have an odd number of elements in all dimensions (e.g., $3\times7$ for a 2D kernel), all kernels in a layer can be stacked along a line passing through their central element (Fig.~\ref{fig:irregular_kernels}).

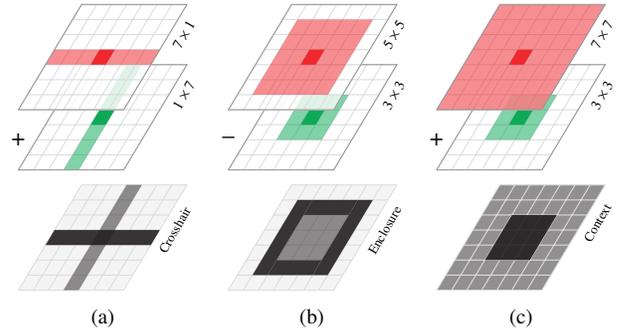
\begin{figure}
	\centering

	\subcaptionbox{\label{subfig:crosshair}}[0.3\columnwidth]
	{
		\begin{tikzpicture}

		\pgfmathsetmacro\xsize{7}
		\pgfmathsetmacro\ysize{7}

		\slantedscope{0}{0.6}{0}{0}{
			\fill[Black,fill opacity=0.05] (0,0) rectangle (\xsize,\ysize);
			\draw[step=1,Black!15] (0,0) grid (\xsize,\ysize);
			\node[draw=none,rotate=60] at (\xsize + 1,4) {\tiny{Crosshair}};
			\fill[Black,fill opacity=0.5] (3,0) rectangle (4,\ysize);
			\fill[Black,fill opacity=0.9] (0,3) rectangle (\xsize,4);
		}
		\kernel{2}{\xsize}{\ysize}{1}{7}{Green}{$1\times7$}
		\kernel{3}{\xsize}{\ysize}{7}{1}{Red}{$7\times1$}

		\node[draw=none] at (0,2) {\small{$+$}};
		\end{tikzpicture}
	}
	\subcaptionbox{\label{subfig:enclosure}}[0.3\columnwidth]
	{
		\begin{tikzpicture}

		\pgfmathsetmacro\xsize{7}
		\pgfmathsetmacro\ysize{7}

		\slantedscope{0}{0.6}{0}{0}{
			\fill[Black,fill opacity=0.05] (0,0) rectangle (\xsize,\ysize);
			\draw[step=1,Black!15] (0,0) grid (\xsize,\ysize);
			\node[draw=none,rotate=60] at (\xsize + 1,4) {\tiny{Enclosure}};
			\fill[Black,fill opacity=0.5] (2,2) rectangle (\xsize-2,\ysize-2);
			\fill[Black,fill opacity=0.9] (1,1) rectangle (2,\ysize-1);
			\fill[Black,fill opacity=0.9] (\xsize - 2,1) rectangle (\xsize - 1,\ysize-1);
			\fill[Black,fill opacity=0.9] (2,1) rectangle (\xsize - 2,2);
			\fill[Black,fill opacity=0.9] (2,\ysize - 1) rectangle (\xsize - 2,\ysize-2);
		}
		\kernel{2}{\xsize}{\ysize}{3}{3}{Green}{$3\times3$}
		\kernel{3}{\xsize}{\ysize}{5}{5}{Red}{$5\times5$}

		\node[draw=none] at (0,2) {\small{$-$}};
		\end{tikzpicture}
	}
	\subcaptionbox{\label{subfig:context}}[0.3\columnwidth]
	{
		\begin{tikzpicture}

		\pgfmathsetmacro\xsize{7}
		\pgfmathsetmacro\ysize{7}

		\slantedscope{0}{0.6}{0}{0}{
			\fill[Black,fill opacity=0.5] (0,0) rectangle (\xsize,\ysize);
			\draw[step=1,Black!15] (0,0) grid (\xsize,\ysize);
			\node[draw=none,rotate=60] at (\xsize + 1,4.5) {\tiny{Context}};
			\fill[Black,fill opacity=0.9] (2,2) rectangle (\xsize-2,\ysize-2);
		}
		\kernel{2}{\xsize}{\ysize}{3}{3}{Green}{$3\times3$}
		\kernel{3}{\xsize}{\ysize}{7}{7}{Red}{$7\times7$}

		\node[draw=none] at (0,2) {\small{$+$}};

		\end{tikzpicture}
	}
	\caption[]{(\subref{subfig:crosshair}) A pair of two long skinny kernels spanning orthogonal dimensions can detect intersecting (crosshair-shaped) features. (\subref{subfig:enclosure}) Detecting enclosures within a single layer is possible by subtracting the activations of two kernels of different shapes. (\subref{subfig:context}) Large kernels can provide contextual information for features extracted by smaller kernels. This information becomes immediately available to the next layer.}
	\label{fig:kernel_combinations}
\end{figure}

The motivation behind this design is that having kernels with different shapes and sizes in the same layer allows the layer to readily `notice' certain features that would otherwise be masked if all kernels had the same shape and size. For instance, long skinny kernels can act as sharp edge detectors without pollution from neighbouring pixels, and pairs of such kernels that extend in different dimensions can act as detectors for crosshair-shaped features (Fig.~\ref{subfig:crosshair}). Furthermore, pairs of kernels that differ by one or more elements in each dimension (for example, $7\times7$ and $5\times5$) would effectively act as enclosure detectors for features that surround the central receptive field shared by the two kernels (Fig.~\ref{subfig:enclosure}). In the same line of thought, large kernels can provide the next layer with contextual information for the finer, sharper features detected by smaller kernels (Fig.~\ref{subfig:context}). In other words, combinations of small and large kernels can reduce the confusion arising from having a lot of fine-grained features obtained from small kernels in the previous layer which cannot be easily associated with each other for lack of longer-range dependency information from one layer to the next.

Viewing the kernel size as a property of each kernel allows us to introduce a new kernel size mutation operator which affects individual kernels rather than the entire layer. Although the implementation of such irregularly shaped layers is fairly straightforward (it requires keeping the kernels as separate tensors and stacking them along the line passing through the central element of each kernel into a single weight tensor before evaluation), none of the existing deep learning libraries support such operations out of the box due to the assumption that the layer has a consistent shape. For example, in the case of $2D$ convolution, a layer is usually implemented as a $4D$ tensor of shape $N\times D\times W\times H$, where $N$ is the batch size, $D$ is the number of input channels, and $W\times H$ represents the size (width and height) of all convolutional kernels. Consequently, it was necessary to use a library which is flexible enough to allow for such a modified evaluation procedure to be implemented while being efficient enough to minimise the impact of the extra stacking operation (cf. \ref{sec:experiments.pycortex}).

Kernels are mutated by either growing or shrinking either the width or height of a single kernel by $2$. For example, a $3\times 5$ kernel could grow into a $5\times 5$ or a $3\times 7$ kernel by adding a $1\times 5$ or a $3\times 1$ block of elements on both sides of the existing kernel along the width or height dimension, respectively. In this way, the functionality of the central part of the kernel is preserved while ensuring that the width and height remain odd so that all kernels in the same layer can be stacked along a line passing through the central element of each kernel. All kernels always span the full depth of the input (i.e., all input channels), and the number of channels does not participate in mutations. Kernels are not allowed to grow larger than half of the input in width or height, and the minimal kernel size is naturally $1\times 1$.

The stride of convolutional layers is evolved by growing or shrinking the stride parameter by $1$ in the width or height dimension. Stride mutations are very disruptive as they do change the output size of the mutated layer and all layers above it, and for that reason stride mutations are much rarer than kernel mutations. The padding parameter is not mutated. Instead, it is set to half of the size of the largest kernel in each dimension (rounded down) to ensure that the size of the response of the convolutional layer is the same as the size of the input regardless of the shapes and sizes of the kernels (cf. Fig. \ref{subfig:conv.k3s1p1}). This is important because otherwise kernel mutations might have a far-reaching effect similar to stride mutations.

\subsection{Crossover}\label{sec:deep_NE.crossover}
The procedure outlined in \ref{sec:cortex} is used for crossover, regardless of whether speciation is enabled or disabled. Crossover proceeds by iterating over matching layers in the two parents and inheriting nodes together with all of their input connections with probability proportional to the parents' relative fitness values. If speciation is disabled and one of the parents happens to contain more layers of a particular type than the other, the excess layers of that type from the larger parent are inherited unaltered with a probability proportional to the fitness of the larger parent. The same principle is applied to nodes in each layer in case of layer size mismatch.

With speciation enabled, crossover is restricted to networks within the same species. At the crossover step, all networks in a species are selected to be parents with probability proportional to their respective relative fitness values (cf. \ref{sec:deep_NE.calibration}). This is done in order to give all networks a chance to reproduce while ensuring that networks with higher fitness would have a better chance to do so. With speciation disabled, any network can participate in crossover with any other network in the ecosystem. In this case, we adopt the following measure for the similarity of the genomes of two networks $n_{i}$ and $n_{j}$:

\begin{equation}
	s_{n_{i},n_{j}} = \frac{O_{i,j}}{N_{i} + N_{j} - O_{i,j}}\label{eq:genotype_similarity}
\end{equation}
where $O_{i,j}$ is the total layer intersection and $N_{i}$ and $N_{j}$ are the total number of nodes in the respective genomes. The total layer intersection $O_{i,j}$ is computed as

\begin{equation}
	O_{i,j} = \sum_{k}{L_{k}(n_{i}) \cap L_{k}(n_{j})} = \sum_{k}{\min(N_{L_{k}(n_{i})}, N_{L_{k}(n_{j})})} \label{eq:genotype_overlap}
\end{equation}

where $L_{k}(n_{i}) \cap L_{k}(n_{j})$ is the intersection of the $k^{th}$ layers in networks $n_{i}$ and $n_{j}$, which amounts to the node count in the smaller of the two layers. This is nonzero only when the layers have the same type (convolutional or FC). For networks with the same number of layers of each type and the same number of nodes in each layer, this similarity measure is $1$, which coincides with the case where speciation is enabled. Since $s_{n_{i},n_{j}}\in [0,1]$, it can be used as a probability for crossover.

\subsection{Culling}\label{sec:deep_NE.culling}
Once the ecosystem grows beyond the preset limit as a result of crossover, a culling procedure takes place to reduce the ecosystem size. New offspring are guaranteed to survive, as well as the champion for each species (or the ecosystem champion with speciation disabled). All other networks are sampled from a weighted distribution with probability weighting $p_{cull}(n_{i})$ for network $n_{i}$ computed as follows:
\begin{equation}
	p_{cull}(n_{i}) = \frac{age(n_{i})}{f_{r}(n_{i})c_{r}(n_{i})} \label{eq:culling}
\end{equation}
where $age(n_{i})$ is the age of network $n_{i}$ in terms of epochs. The culling continues by selecting networks one by one until the ecosystem size limit is reached.

\subsection{PyCortex NE platform}\label{sec:experiments.pycortex}
We developed a NE platform (PyCortex) which implements the above framework, including a mutation operation capable of altering the size of individual kernels as outlined in \ref{sec:deep_NE.mutation.conv_layers} as well as other common mutation operations (adding and removing nodes and entire layers and mutating the stride of convolutional layers). In essence, PyCortex provides a convenient interface to an established deep learning platform (PyTorch\footnote{PyTorch was chosen for its flexible interface which allows network evaluation graphs to be generated at runtime, which was essential for the efficient implementation of convolutional layers with varying kernel sizes.}) and can be used for direct evolution of both regular and convolutional deep NNs by abstracting the computational details of the crossover and mutation operations. Each evolved network is a valid model which can be evaluated directly in PyTorch, harnessing all the effort that has been invested into making the platform efficient and flexible. At present, PyCortex employs a hybrid strategy where the network structure is evolved while weights are optimised with BP. However, PyTorch provides easy access to all learnable parameters in a model, which paves the way to testing alternative weight optimisation algorithms, such as evolutionary strategies. The proposed platform is released as an open-source project\footnote{\url{https://gitlab.com/cantordust/pycortex}.} to facilitate research in NE. We hope that as it matures it can serve as a standard tool for prototyping and benchmarking novel NE algorithms.

\section{Experiments}\label{sec:experiments}
We conducted experiments on several image classification tasks to test the feasibility of the proposed framework with respect to deep convolutional models. Following the insight in \parencite{SpringenbergDosovitskiyBroxRiedmiller--2014}, the layer types were limited to convolutional and FC, without pooling layers. All experiments were performed with and without speciation to test whether crossover in the case without speciation would result in lower offspring fitness (cf. Fig. \ref{fig:layered_model_crossover}). All other configuration options were identical across the experiments. With speciation enabled, ecosystems were initialised with $8$ species, and a hard limit of $16$ species was used in all experiments.

The experiments were conducted with initial and maximal ecosystem size of $64$ and $111$, respectively\footnote{The number $111$ was an artefact of the cluster configuration. Each node on the particular cluster we used contained $28$ cores, and four nodes were used for each experiment, raising the total core count to $112$. However, one core was reserved for the master MPI process, bringing the total number of available cores to $111$. This number was used as the maximal ecosystem size to ensure that each network had a dedicated CPU core for evaluation.}. Connection weights were drawn from a normal distribution with mean $0$ and standard deviation of $0.1$, both during ecosystem initialisation and when any new connections were added through mutations.

Two important points about the training procedure are worth emphasising.

\begin{itemize}
	\item{At each generation (training epoch), each network in the ecosystem was trained on a \textbf{random 10\% subset} of the training data, after which it was evaluated on the entire test set. The accuracy on the test set was used as the absolute fitness of that network. The decision to use a small random portion of the data for training was motivated by the aim to examine whether crossover can transfer learned features to the offspring in an epigenetic evolution scenario. For that purpose, all new offspring were evaluated on the test set prior to commencing training with BP.}

	\item{We used a relatively large training batch size of $128$. Large training batches tend to produce solutions which converge to `sharp' minima \parencite{KeskarMudigereNocedalSmelyanskiyTang--2016}, with a negative impact on generalisation. Hence, it was of particular interest to examine whether the perturbations introduced by the mutation and crossover operations would still allow the networks to generalise well even with a large batch size.}
\end{itemize}

\subsection{Datasets}\label{sec:experiments.datasets}
Four commonly used datasets (MNIST \parencite{LeCunCortes--2010}, SVHN \parencite{NetzerWangCoatesBissaccoWuEtAl--2011}, Fashion-MNIST \parencite{XiaoRasulVollgraf--2017}) and CIFAR-10 \parencite{KrizhevskyHinton--2009}) were used in the experiments. All datasets have $10$ classes. MNIST and SVHN contain images of digits (preprocessed and grayscale in the case of MNIST, natural RGB in the case of SVHN), while Fashion-MNIST and CIFAR-10 contain images of $10$ different types of objects (preprocessed and grayscale in the case of Fashion-MNIST, natural RGB in the case of CIFAR-10). This allowed us to compare potential differences arising from the use of colour information. Each experiment was run $10$ times for $100$ generations per run.

\subsection{Results}\label{sec:results}
\begin{figure}
	\centering
	\subcaptionbox{\label{subfig:mnist.with_spec.highest_fitness}}[\linewidth]
	{
		\includegraphics[width=\linewidth]{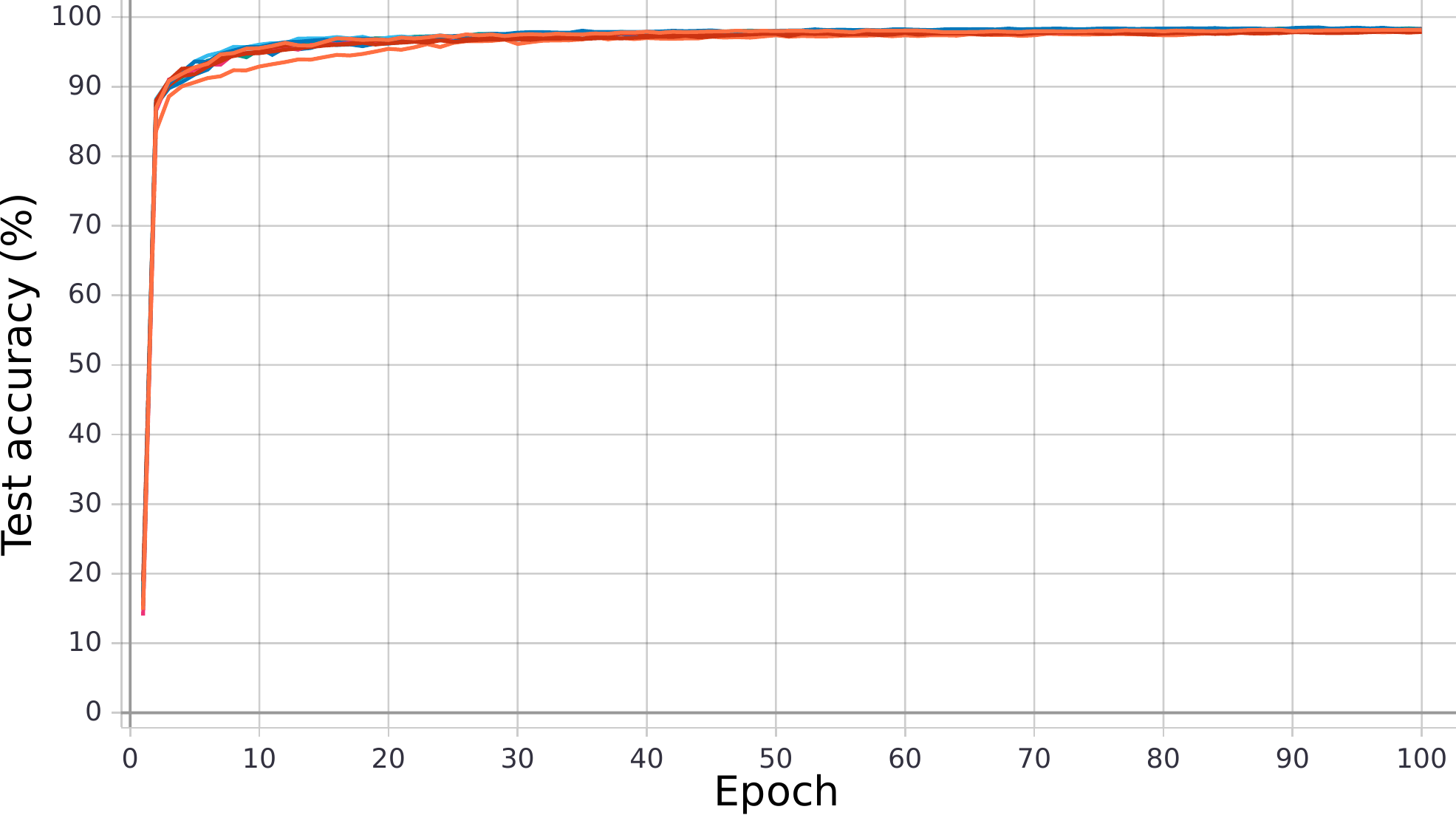}
	}
	\subcaptionbox{\label{subfig:mnist.with_spec.average_fitness}}[\linewidth]
	{
		\includegraphics[width=\linewidth]{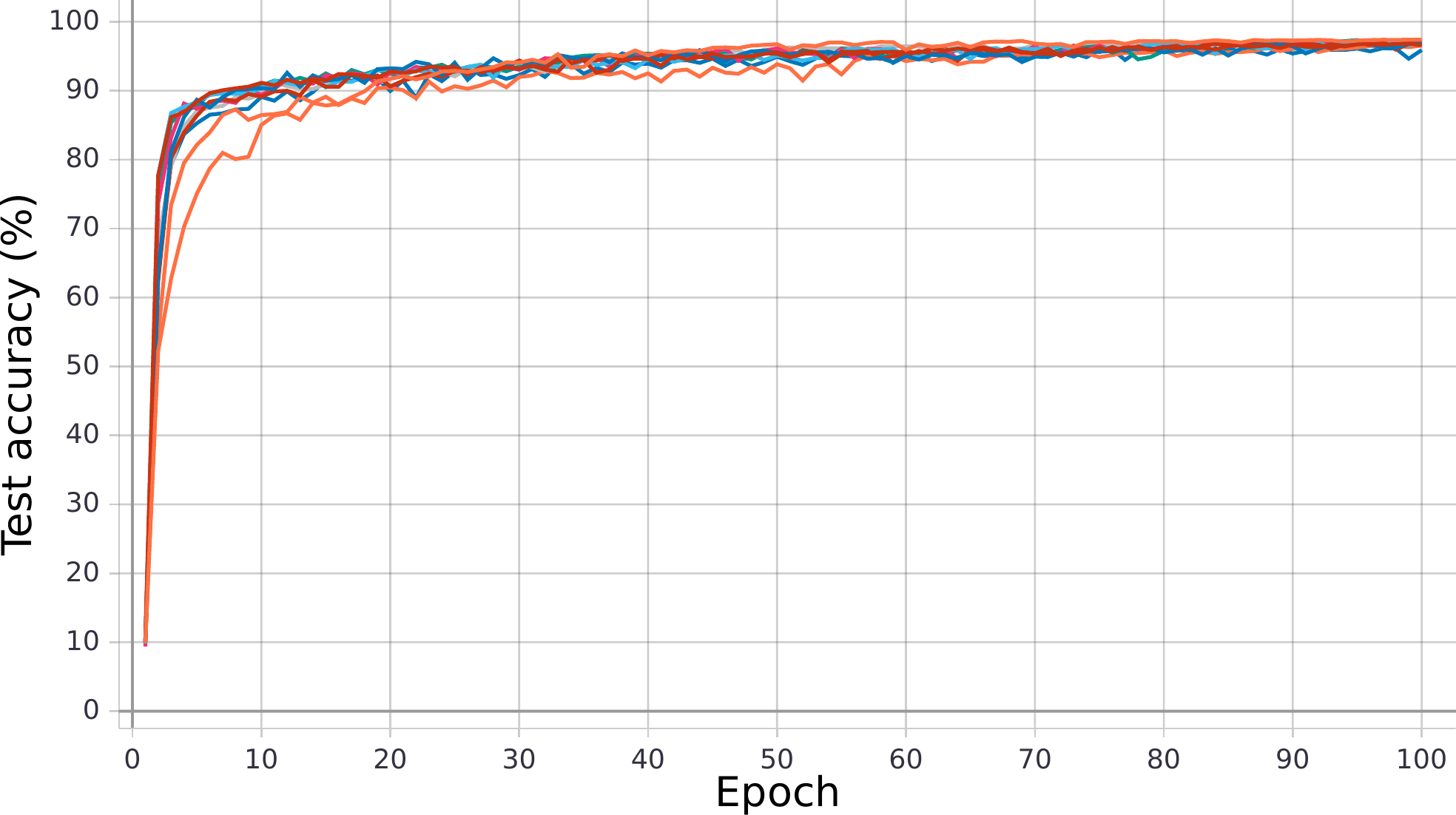}
	}
	\subcaptionbox{\label{subfig:mnist.with_spec.avg_offspring_fitness_before_BP}}[\linewidth]
	{
		\includegraphics[width=\linewidth]{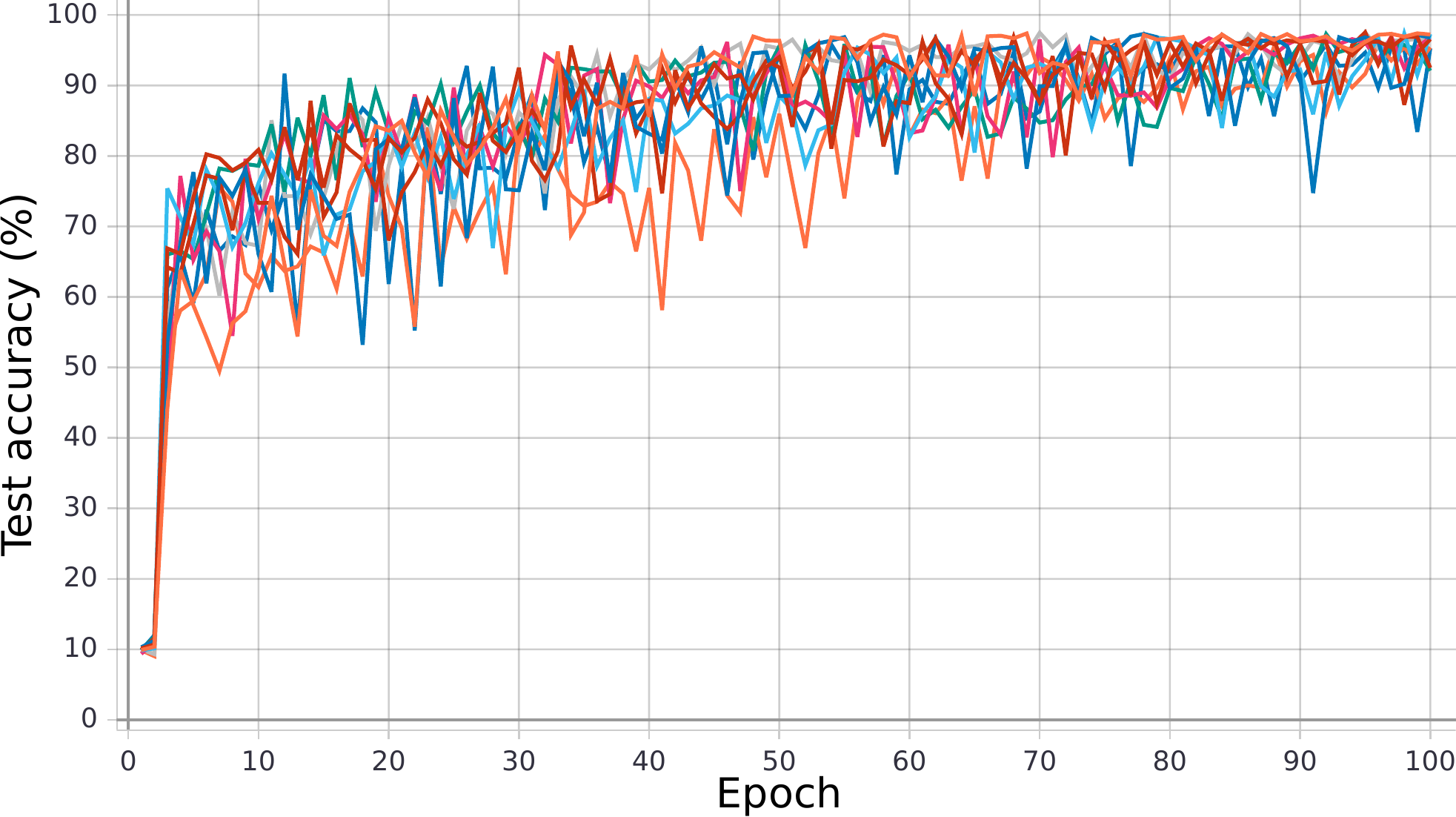}
	}
	\caption[]{Experiment results for the MNIST benchmark with speciation enabled. (\subref{subfig:mnist.with_spec.highest_fitness}) Highest fitness, (\subref{subfig:mnist.with_spec.average_fitness}) average fitness and (\subref{subfig:mnist.with_spec.avg_offspring_fitness_before_BP}) average fitness of new offspring before BP.}
	\label{fig:results_mnist.with_spec}
\end{figure}

One of the key advantages of using an established platform for evaluation is that all the tools available for that platform can be readily used to track and analyse the learning progress and other parameters. Specifically, we used the tensorboardX module for saving epoch data in TensorBoard log format directly from PyTorch, which has the added advantage of being able to plot the data in real time. The results for the highest fitness, average fitness and average offspring fitness before BP for all experiments are summarised in Table~\ref{table:experiments.results} (results for MNIST with speciation enabled are presented in Fig.~\ref{fig:results_mnist.with_spec}).

\begin{table}
	\caption[]{Highest value, mean and standard deviation recorded over $10$ runs after epoch $100$ for the highest fitness, average fitness and average offspring fitness before BP for each experiment.}
	\begin{center}
		\begin{tabular}{|c|c|c|c|}
			\hline
			                       Experiment                        &       \thead{Highest\\fitness}        &      \thead{Average\\fitness}       & \thead{Average offspring\\fitness before BP} \\
			\hline
			   \thead{MNIST \\ \footnotesize{(with speciation)}}     & \thead{98.32\% \\ (98.13 $\pm$ 0.15)} & \thead{97.14\% \\ (96.8 $\pm$ 0.4)} &     \thead{97.28\% \\ (95.6 $\pm$ 1.8)}      \\
			\hline
			    \thead{MNIST \\ \footnotesize{(no speciation)}}      & \thead{98.14\% \\ (97.98 $\pm$ 0.13)} & \thead{95.31\% \\ (93.7 $\pm$ 1.1)} &     \thead{82.98\% \\ (72.5 $\pm$10.7)}      \\
			\hline
			    \thead{SVHN \\ \footnotesize{(with speciation)}}     & \thead{82.84\% \\ (79.86 $\pm$ 2.06)} & \thead{76.9\% \\ (70.2 $\pm$ 7.5)}  &     \thead{77.25\% \\ (66.5 $\pm$ 8.0)}      \\
			\hline
			     \thead{SVHN \\ \footnotesize{(no speciation)}}      & \thead{81.20\% \\ (80.22 $\pm$ 0.89)} & \thead{70.62\% \\ (66.5 $\pm$ 3.1)} &     \thead{55.53\% \\ (44.4 $\pm$ 7.4)}      \\
			\hline
			\thead{Fashion-MNIST \\ \footnotesize{(with speciation)}} & \thead{89.49\% \\ (88.54 $\pm$ 0.70)} & \thead{87.89\% \\ (85.3 $\pm$ 1.4)} &     \thead{88.18\% \\ (82.0 $\pm$ 4.3)}      \\
			\hline
			 \thead{Fashion-MNIST \\ \footnotesize{(no speciation)}}  & \thead{88.88\% \\ (88.44 $\pm$ 0.36)} & \thead{85.53\% \\ (83.3 $\pm$ 1.1)} &     \thead{83.61\% \\ (64.9 $\pm$ 7.8)}      \\
			\hline
			  \thead{CIFAR-10 \\ \footnotesize{(with speciation)}}   & \thead{55.52\% \\ (51.35 $\pm$ 3.44)} & \thead{46.45\% \\ (43.9 $\pm$ 4.8)} &     \thead{45.94\% \\ (42.7 $\pm$ 5.1)}      \\
			\hline
			   \thead{CIFAR-10 \\ \footnotesize{(no speciation)}}    & \thead{54.4\% \\ (52.44 $\pm$ 0.85)}  & \thead{43.87\% \\ (41.9 $\pm$ 1.1)} &      \thead{35.8\% \\ (32.4 $\pm$ 1.8)}      \\
			\hline
		\end{tabular}
		\label{table:experiments.results}
	\end{center}
\vspace{-0.5em}
\end{table}

\section{Discussion}\label{sec:discussion}
The evolved CNNs performed admirably on the MNIST dataset, reaching an overall high scores of $98.32\%$ and $98.14\%$ with and without speciation, respectively, whereas the highest score for the SVHN dataset was considerably lower (although still above $80\%$). This reflects the higher difficulty of classifying natural images versus pre-processed ones, as the objective is the same in both cases (classifying digits). The same trend can be observed in the case of Fashion-MNIST and CIFAR-10, where it is even more pronounced.

The results in Table~\ref{table:experiments.results} reveal surprisingly small differences for the highest recorded fitness between the experiments with and without speciation. In previous research, we employed speciation to protect mutated networks from being eliminated from the ecosystem before they had had a chance to optimise any new weights introduced by mutations. However, in that case weights were optimised by evolution, whereas in the above experiments the weights were optimised by BP, which likely reduces the importance of speciation for this particular purpose. Furthermore, the experiment logs (not shown) revealed that in the experiments with speciation disabled the parameter count increased steadily over the $100$ epochs, whereas in those with speciation enabled it was essentially stagnant and even \textit{decreased} in a number of runs. In all cases with speciation enabled, the logs also revealed a large number of failed structural mutations (addition and removal of nodes and layers) due to reaching the limit on the species count, which is likely the cause for the stagnation. In this regard, the architectures evolved in all experiments were relatively small (\textasciitilde$10^{5}$ parameters), which can at least partially explain the low scores obtained on CIFAR-10 and, to a smaller extent, SVHN and Fashion-MNIST. Nevertheless, for all four datasets, the average overall fitness and the average offspring fitness before BP at epoch $100$ are higher in the cases with speciation than in those without, but more experiments are necessary in order to run a proper statistical analysis on the significance level of this difference. In future work, we plan to develop a more robust method for determining the mutation rate for complexifying mutations in order to evolve much larger models with potentially millions of parameters. We are also working on a dynamic speciation limit which is designed to altogether eliminate the need to set a hard species limit, allowing the number of parameters to increase more rapidly when speciation is enabled.

Perhaps the most satisfying part of the results is the average offspring fitness before BP, which represents the performance of new offspring evaluated \textit{before any training}. As outlined in \ref{sec:experiments}, all networks were trained on a random $10\%$ subset of the training data at each epoch, which means that no single network ever saw the entire training dataset. However, over the course of $100$ epochs, practically all of the dataset would have been \textit{collectively} seen by the ecosystem. The only way that this could be useful to new offspring would be if crossover could successfully preserve and transfer learned features by combining useful genes (nodes with all of their input connections) from the parents into the offspring. The results for the average offspring fitness before BP, which increased steadily over the $100$ epochs in all cases (cf. Fig. \ref{subfig:mnist.with_spec.avg_offspring_fitness_before_BP}; a similar trend was observed in all other experiments), provides direct confirmation that this is in fact the case. There is also a clear difference in the results for the offspring fitness before BP between experiments with and without speciation, which confirms the intuition presented in \ref{sec:cortex} that speciation affects the fitness of new offspring by determining the way genes and chromosomes are matched during crossover.

\section{Conclusion}\label{sec:conclusion}
This study extended previous research on NE and demonstrated its applicability to the direct evolution of deep convolutional models. A new convolutional layer layout which allows kernels of different size and shape to coexist within the same convolutional layer was also proposed, and a corresponding mutation operator which can resize individual kernels was introduced. A detailed analysis and visualisation of the kernels in evolved CNNs will be presented in future work. Furthermore, the crossover procedure previously proposed for unstructured networks was extended to deep layered networks, and its feasibility was demonstrated through image classification experiments with evolved CNNs. Finally, a NE platform which implements the proposed framework, including crossover and kernel mutation operators for deep CNNs, was developed on top of an established deep learning library (PyTorch). This platform is released as an open-source project with the aim to provide a common framework for prototyping, evaluating and comparing NE algorithms.

\section*{Acknowledgment}
The authors would like to thank the anonymous reviewers for their insightful comments and suggestions, which helped improve the readability and the overall quality of the paper. This research was undertaken with the assistance of resources and services provided by the National Computational Infrastructure, which is supported by the Australian Government.

\printbibliography

\end{document}